\documentclass[12pt]{article}

\usepackage{mdwlist}
\usepackage{graphicx}
\usepackage[cmex10]{amsmath}
\usepackage{amsfonts}
\usepackage{amssymb}
\usepackage{algorithmic}
\usepackage{algorithm}
\usepackage{ntheorem}
\usepackage{url}
\begin{document}

%\begin{frontmatter}



%\dochead{Computer Networks}

\title{Measuring the Complexity of \\Ultra-Large-Scale Adaptive Systems}


\author{Michele~Amoretti$^1$, Carlos Gershenson$^2$\\
$^1$ Centro Interdipartimentale SITEIA.PARMA, \\ 
Universit\`a degli Studi di Parma, Italy\\
Parco Area delle Scienze 181a, 43124 Parma, Italy\\
email: michele.amoretti@unipr.it\\
phone: +39 0521 906147\\
fax: +39 0521 905723\\
$^2$ Computer Science Department, IIMAS,\\ Universidad Nacional Aut\'onoma de M\'exico}

\maketitle 
\begin{abstract}
Ultra-large scale (ULS) systems are becoming pervasive. They are inherently complex, which makes their design and control a challenge for traditional methods. Here we propose the design and analysis of ULS systems using measures of complexity, emergence, self-organization, and homeostasis based on information theory. These measures allow the evaluation of ULS systems and thus can be used to guide their design. We evaluate the proposal with a ULS computing system provided with adaptation mechanisms. We show the evolution of the system with stable and also changing workload, using different fitness functions. When the adaptive plan forces the system to converge to a predefined performance level, the nodes may result in highly unstable configurations, that correspond to a high variance in time of the measured complexity. Conversely, if the adaptive plan is less "aggressive", the system may be more stable, but the optimal performance may not be achieved. 
\end{abstract}

%\begin{keyword}
%ultra-large-scale system \sep peer-to-peer \sep evolution \sep complexity \sep information theory


%\end{keyword}

%\end{frontmatter}

% \linenumbers


\section{Introduction}
Ultra-large-scale (ULS) systems \cite{SEI06} are the result of the interconnection of heterogeneous IT systems --- characterized by decentralized goals and control --- that as a whole exhibit one or more properties (\textit{i.e.} behavior) which are not easily inferred from the properties of the individual parts. ULS are complex systems, since the interactions of their components determine their future state and that of the system \cite{GershensonHeylighen2005}. This interconnectedness limits the predictability of ULS, since interactions generate novel information, thus making traditional methods that rely on prediction inadequate \cite{GershensonFoS}.

Examples of ULS systems are the Internet, healthcare infrastructures, e-markets, global ambient intelligence systems, and distributed high-performance computing facilities.  
To overcome the rapidly growing complexity of their management, and to reduce the barrier that complexity poses to further growth, a variety of architectural frameworks based on ``self-regulating" components has been proposed \cite{GershensonDCSOS,Psaier2011}. Adaptive plans may turn ULS systems into a more efficient, environment-driven systems, provided that the adaptive plan itself does not introduce further complexity and instabilities \cite{DSN10}.

Even when there are several examples of ULS and the importance of their adaptivity is well accepted, there is no standard method for developing adaptive ULS. Even more, there is no framework to measure how adaptive different ULSs are. In this paper, we propose the evaluation of adaptive ULS using complexity measurement principles introduced by Gershenson and Fernandez in \cite{Gershenson2012}. We apply these to a peer-to-peer system designed with the Distributed Remodeling Framework (DRF) defined by Amoretti in \cite{Amoretti2013}. Our contribution lies in the proposal of a methodology to measure the adaptivity of any ULS, facilitating their comparison. 

To the best of our knowledge, this is the first proposal of a general methodology for measuring the complexity of ULS systems, independent of the particular implementation. Usually, benchmarks are adopted to test the characterize single parts of ULS systems (which are always built with a bottom-up approach), and partial indicators are used to analyze the whole system in simulation \textit{et al.} \cite{Foo2011}.

The paper is organized as follows: In the following section, related work concerning ULS evolutionary systems is mentioned. In section \ref{methodology}, the proposed methodology is presented. In section \ref{casestudy}, the case study of an ULS peer-to-peer computing system with distributed genetic adaptation is used as an illustration. In section \ref{simulations}, results of the case study evaluation carried out by means of discrete event simulation are shown. Section \ref{conclusion} concludes the paper with a summary of achieved results and an outline of future work.

\section{Related Work}
\label{relatedwork}

\subsection{ULS Systems}

The SEI study \cite{SEI06} on ULS systems brings together experts in software and other fields to examine
the consequences of rapidly increasing scale in software-reliant systems. The report details a broad,
multi-disciplinary research agenda for developing the ultra-large-scale systems of the future, that also include 
computer-supported evolution, adaptable structure and emergent qualities. 

All these aspects can be placed under the umbrella of autonomic computing, that proposes 
to provide distributed systems with four key properties: self-configuration, self-healing, self-optimization, and
self-protection \cite{IBM03}. IBM has suggested a reference model for autonomic control loops, which is sometimes called the MAPE-K (Monitor, Analyze, Plan, Execute, Knowledge) loop \cite{HM08}. This model is being widely used to communicate the architectural aspects of autonomic systems. The MAPE-K construction can be iterated, as the control loop itself could be adaptive (see for example \cite{MullerSchloer2011} and \cite{Bruni2012}).

Nakano and Suda have proposed a distributed approach for evolutionary adaptation of network services \cite{Nakano2005}, based on agents that migrate, replicate, reproduce and die. An agent invokes its behavior based on its internal state and environmental conditions, using a weighted sum aggregation method. For example, when the weighted sum for reproduction exceeds its reproduction threshold, the agent selects a reproduction partner from agents that are within $n$ hops.

Hales introduced an algorithm called SLAC, which means selfish link and behavior adaptation
to produce cooperation \cite{H04}. SLAC is based on the copy and rewire approach, whose basic algorithm
assumes that peer nodes have the freedom to change the way they handle and dispatch requests to
and from other nodes, and drop and make links to nodes they know about. Another interesting approach
for node restructuring has been proposed by Tyson \textit{et al.} \cite{TGM08}, that show how survival of the fittest
has been implemented into the Juno middleware. On receipt of a superior component, Juno dynamically
reconfigures the internal architecture of the node, by replacing the existing component with the new one.

\subsection{Peer-to-Peer Networks}

Distributed control is the main reason for the intrinsic \textit{robustness} and \textit{resiliency} of complex adaptive systems. Actually, they are relatively insensitive to perturbations or errors, and have a strong capacity to restore themselves. Due to their distributed organization, the non-damaged regions can usually make up for the damaged ones.

Distributed control is also a characterizing aspect of peer-to-peer (P2P) networks \cite{Amoretti2009, Cooper2006, Karakaya2008, Esposito2011, Chen2012}, where every node (the peer) acts as server and client, by collecting, providing and consuming resources (such as CPU cycles, storage, and bandwidth). A P2P network is usually an overlay network which builds over existing physical infrastructures. Some overlay schemes (such as those of the DKS family \cite{Alima2003}) lead to high scalability, efficient resource sharing mechanisms and robustness to node failure. 
The drawbacks are that administration may be difficult (one person cannot determine the whole accessibility setting of the whole network), malicious software can be easily transmitted, and in general both normal functioning and recovery protocols must be based on distributed algorithms, by taking into account that emergent behaviors are difficult to predict.

P2P networks are complex, but distributed control is not sufficient for considering them adaptive. Usually, the peers have a static structure and execute prefixed protocols. 
To improve the efficiency of peer-to-peer systems while adapting to changing environmental conditions, static peer-to-peer protocols can be replaced by \textit{adaptive plans}. Indeed, the introduction of artificial evolution in peer-to-peer networks allows to face environmental needs preserving high performance, while saving resources in a totally unsupervised manner. In a seminal paper on this topic, Paechter \textit{et al.} call this concept \textit{infoworld evolution} \cite{Paechter2000}.

Evaluating the complexity of distributed adaptive systems, such as P2P networks, is a challenging task. Even more challenging is to exploit complexity measurements for improving the design of such systems.

\subsection{Complexity Measurement}

There are dozens of measures of complexity. Several of them are based on information theory \cite{Prokopenko2009}. This is convenient because anything can be measured in terms of information, thus these measures can be applied to any studied phenomenon. Some measures of complexity correlate with ``disorder" or ``chaos", making random phenomena to have the highest complexity. However, other approaches consider complexity as a balance between chaos and order. This balance is desirable for computing and living systems, since they require certain stability but also certain variability. The ordered extreme is robust, but does not enable adaptation. The chaotic extreme allows for adaptation and exploration, but information cannot be lost. The best possible adaptation can only occur in the critical regime between ordered and chaotic dynamics~\cite{Langton1990,Kauffman1993}.
In this approach, complexity is seen as a balance between propagating and transforming information \cite{Gershenson2007}. Thus, systems with a higher complexity will be more adaptable than those which are not.

\section{Methodology}
\label{methodology}
The ULS systems we want to characterize are made of several thousands nodes that interact in a peer-to-peer fashion, \textit{i.e.} without a centralized control.
To describe the functional architecture and behavior of such nodes, we use the Distributed Remodeling Framework (DRF) \cite{Amoretti2013}.
Each peer has a modular structure that can dynamically change, by adding or removing components. 
Being $p$ the total number of component types for the system, we define a vector $\mathbf{M}$ of $q \leq p$ components, called the \textit{model} of the peer. An adaptive plan $\tau$ produces a sequence of configurations, \textit{i.e.} a trajectory through the \textit{search space} $\mathcal{M}$ of models, following an evolutionary process. 

By means of the framework introduced by Gershenson and Fernandez in \cite{Gershenson2012}, we measure the evolution over time of the information $I$ associated to $\mathbf{M}$, when the components of the latter are integer values $x \in [1..n]$ used as parameters for the node's functional processes. Taking into account the latest $W$ configurations of $\mathbf{M}$, we measure the frequency of each value for $x \in [1..n]$. Then we use the frequency as $P(x)$ and compute the normalized information as

\begin{equation}
I = \frac{- \sum P(x)\log_2 P(x)}{I_{max}} 
\end{equation}

where $I \in [0,1]$ and $I_{max} = -\log_2 1/n$, since the maximum information value is achieved when all values $1,..,n$ have the same probability \cite{Gershenson2012}. Minimum information ($I=0$) occurs when only one value is repeated in time. This implies that the node is static, i.e. there is no change.

For example, if the latest $W=3$ configurations of $\mathbf{M}$ have been $(1,3,5)$, $(1,3,6)$, $(1,4,6)$, then the frequencies are $P(1) = 1/3$, $P(2) = 0$, $P(3)=P(6)=2/9$, $P(4)=P(5)=1/9$. The associated information is 
\begin{equation*}
I = \frac{-1/3\log_2 1/3 -4/9\log_2 2/9 -2/9\log_2 1/9}{-\log_2 1/6} = 0.85
\end{equation*}

Since peers are randomly initialized, we assume a random ``input" and use the following simplified formulas \cite{Gershenson2012}:

\begin{equation}
\text{emergence} \quad E = I
\end{equation}

Higher emergence implies that the process \emph{produces} more information. Low emergence implies that the process does not produce information, i.e. $I=0$.

\begin{equation}
\text{self-organization} \quad S = 1- I
\end{equation}

In this simplified equation, self-organization can be seen as the opposite of emergence, as organization can be seen as a reduction of entropy~\cite{GershensonHeylighen2003a}, which is equivalent to $I$. Low emergence implies high self-organization and vice versa. The most self-organized process is that which is static, while the least self-organized process is that which changes the most, i.e. it is the most emergent ($I=1$).

\begin{equation}
\text{complexity} \quad C = 4 \cdot E \cdot S
\end{equation}

where the $4$ factor is for normalization reasons, since $I \in [0,1]$ and $C$ is maximized when $I = E = S = 0.5$. Complexity here is seen as a balance between emergence (change) and self-organization (order). Complexity is low when $E$ or $S$ are high, while it is maximal when they are equal. 

\begin{equation}
\text{homeostasis} \quad H = 1 - d(I, I_{init})
\end{equation}

where $d$ is the normalized Hamming distance between the $\mathbf{M}$ associated to $I$ and $I_{init}$, which is the information associated to the initial configuration of the peer. $H$ is a complementary measure of change in the system. The highest $H=1$ is given when there is no change, $H \approx 1/n $ indicates lack of correlation between compared states. 

A high $E$, $S$ or $H$ implies low adaptivity, since a system with high $E$ implies constant change (close to noise), while high $S$ and $H$ imply a static system, which cannot adapt to changes in its demand. A high $C$ implies adaptivity of the system.

\section{Case study}
\label{casestudy}
We use the example of an ultra-large-scale computing system where nodes form a peer-to-peer overlay network, whose global functioning is the result of the interactions among nodes, and depends on their local evolution based on an adaptive plan implemented as a genetic algorithm (GA). Such a case study is particularly interesting because of the self-adaptation properties of the considered system. Our analysis shows that adaptive plans, being more or less "aggressive", can lead to very different results. The design of such systems requires to find a good tradeoff between stability and optimality of performance. Such a challenging task can be effectively supported by the complexity measurement methodology defined above.

In the considered system, shared resources cannot be acquired by replication once discovered, but may only be directly used upon contracting with their hosts. An example of such resources is disk space, which can be partitioned and allocated to requestors for the duration of a task, or in general for an arranged time. 

The topology of the envisioned ultra-large-scale system can be represented as an undirected graph whose arcs stand for mutual knowledge among peers. The node degree $k$ is the number of neighbors of each node. Its statistical distribution depends on the history and on the dynamics of the network, and may affect the performance of the distributed algorithms which are executed.

Every peer executes Algorithm \ref{alg:lookup}, which is based on \textit{epidemic} query propagation \cite{Eugster04}, to search for resources. 
Lookup messages are matched with local resources and, if necessary, propagated to neighbor peers. 
Every peer is provided with a cache that contains the descriptors of the remote resources that have been previously discovered. 
Such a cache is inspected before sending random (\textit{i.e.} blind) queries. Lookup messages have a time-to-live ($T$), which represents the remaining number of hops before expiration. Every peer caches received queries, in order to drop subsequent duplicates. The reason of such an approach for resource lookup is that bio-inspired epidemic protocols are scalable and robust against network failures, while providing probabilistic reliability guarantees \cite{Ahi07}.

\begin{algorithm}
\caption{Resource lookup}
\label{alg:lookup}
\begin{algorithmic}[1]
\begin{footnotesize}
\IF {resource is locally available}
	\STATE notify interested peer
	\IF {interested peer still interested}
		\STATE provide resource instance to interested peer
	\ENDIF
\ELSE
	\STATE $T \leftarrow T-1$
	\IF {resource descriptor is in cache}
		\STATE propagate query to cached peer
	\ELSE
		\IF {$T > 0$}
			\STATE propagate query to (all or some) neighbors
		\ENDIF
	\ENDIF
\ENDIF
\end{footnotesize}
\end{algorithmic}
\end{algorithm}

Resource lookup is affected by the following parameters:
\begin{itemize}
\item $f_k =$ fraction of neighbors targeted for query propagation
\item $T_{max} =$ propagations depth, \textit{i.e.} the maximum number of times a query is forwarded by the peers before being removed from the network
\item $D_{max} =$ maximum size of the cache
\end{itemize}
that contribute to the definition of the peer's phenotype:
\begin{equation}
\label{eq:Phi}
\Phi \doteq \phi_0f_k + \phi_1T_{max} + \phi_2D_{max}
\end{equation}
In the previous formula, the constant weights $\phi_0,\phi_1,\phi_2 \in \mathbb{R}$ control the influence of each parameter. 
Such weights are set in advance to define the priorities of the three components of $\Phi$. 
Of course one could set them to $1$ and let such priorities emerge from the evolution of the genotype (defined below).

In traditional epidemic algorithms, the parameters have fixed values, and are set in advance according to the results of some tuning process, which should guarantee good performance with high probability. 
Actually, if all peers were configured with $f_k = 1$ and the same $T$ value, the resulting scheme would match the Gnutella protocol \cite{Gnut}. The purpose of introducing artificial evolution of these peers is to guarantee high system performance, while avoiding overabundant resource lookup messaging within the network. 
In the adaptive approach, the three parameters listed above are made functions of the chromosome, and are randomly initialized when a peer is created and joins the network.
Moreover, such parameters are adaptively tuned, according to a fitness function $F$ that is to be minimized, taking into account the average \textit{query hit ratio}:
\begin{equation}
\label{eq:QHR}
\langle QHR \rangle = AQ(QHR_0,..,QHR_k) = \frac{1}{k+1} \sum_{j = 0}^k {QHR}_j
\end{equation}
The query hit ratio is defined as the number of query hits $QH$, \textit{i.e.} successful lookups, versus the number of submitted queries $Q$.
In the previous equation, it is assumed that $j = 0$ for the considered peer, and $j = 1,..,k$ for its neighbors.

An increasing request rate penalizes the lookup performance, because the shared resources (\textit{e.g.} CPU, RAM, disk space) are consumable. 
Thus, the fitness function must drive the peer in adapting its lookup parameters to the rate of direct or indirect user requests. 
Each peer applies the adaptive plan to its current model, and evolution can be either autonomous or based on the interaction with the peers' neighbors. 
In this case study, the second approach was adopted: each peer evolves its model by taking into account its neighbors' models (see below Algorithm \ref{alg:adaptation}).

Every peer is characterized by a model $\mathbf{M}$ (genotype), defined by three genes $\{M_0, M_1, M_2\}$  with values in a limited subset of $\mathbb{N}$. The following equations define the mapping between the phenotype and genotype:
\begin{itemize}
\item $f_k = c_0 M_0$
\item $T_{max} = c_1 M_1$
\item $D_{max} = c_2 M_2$
\end{itemize}
where $c_i \in \mathbb{R}$, (with $i=0,1,2$) are constants.

\subsection{Fitness functions}

Examples of suitable fitness functions are:
\begin{alignat}{2}  
F_1(\Phi, \langle QHR \rangle) &= \begin{cases}
       1 / \Phi  \quad \text{if $\langle QHR \rangle < \beta$} \cr
       \Phi \quad \text{if $\langle QHR \rangle > \beta$}
       \end{cases} \nonumber \\
F_2(\Phi, \langle QHR \rangle) &= (1 - \langle QHR \rangle)\frac{1}{\Phi} + \langle QHR \rangle \Phi \nonumber \\
F_3(\Phi, \langle QHR \rangle) &= (\frac{1}{\langle QHR \rangle + \delta} - 1)\frac{1}{\Phi} + \langle QHR \rangle \Phi \nonumber \\
F_4(\Phi,\langle QHR \rangle) &= ( \frac{\langle QHR \rangle}{\beta} - 1) (\frac{\Phi}{\Phi_M}) + (1- \langle QHR \rangle) \nonumber
\end{alignat}
where $\beta$ is a threshold value, $\delta$ is a small parameter for avoiding division by zero when $\langle QHR \rangle = 0$, and $\Phi_M = max\{\Phi\}$.

These fitness functions favor lower phenotypes when the average query hit ratio $\langle QHR \rangle$ is high, and higher ones when the $\langle QHR \rangle$ is low. If the values of the parameters which define the phenotype increase, the search process becomes more expensive in terms of time/space, but at the same time the probability of successful lookup (approximated by $QHR$) should be improved. 

\begin{figure*}[!hbt]
\centering
\includegraphics[width=8cm]{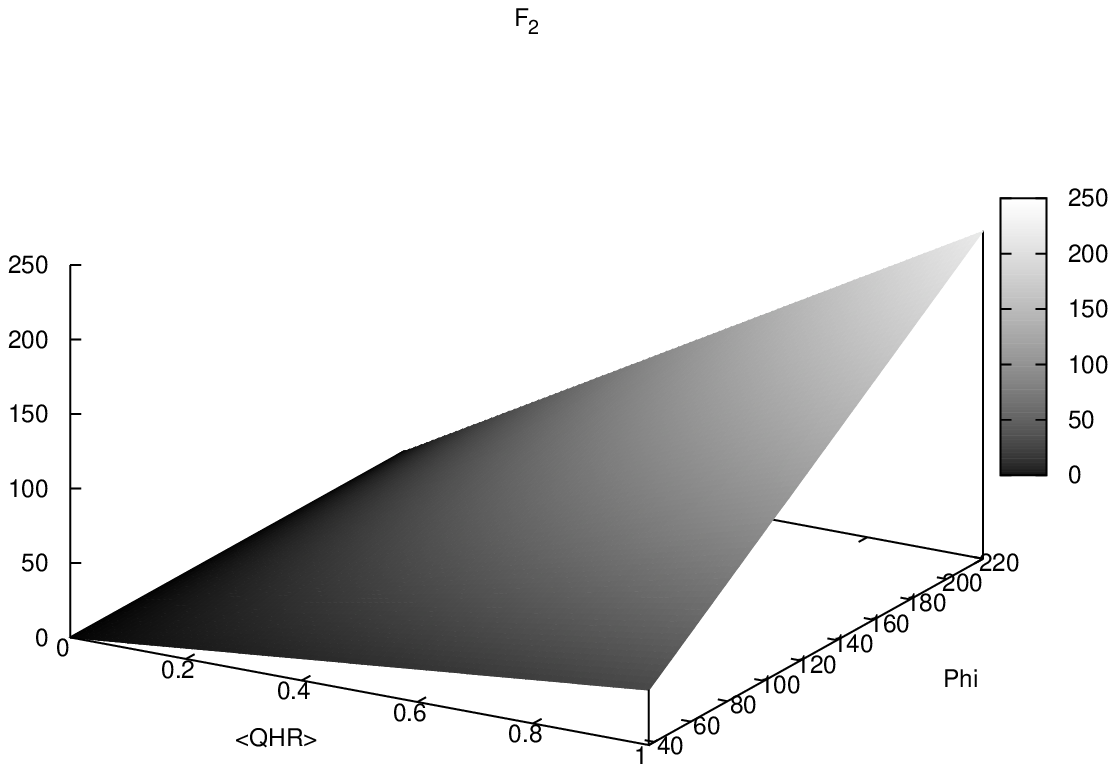}
\includegraphics[width=8cm]{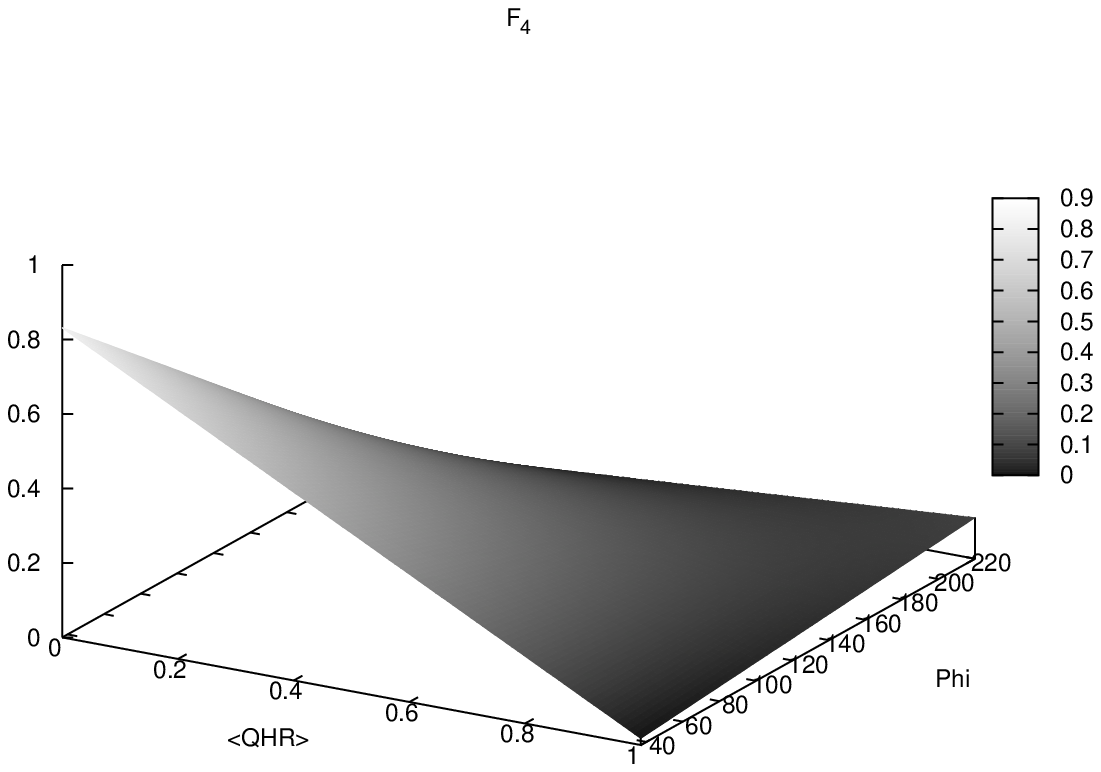}
\caption{Plot of $F_2$ and $F_4$ with the parameters' values of the simulation experiments.}
\label{fig:Fi}
\end{figure*}

$F_1$ and $F_4$ allows to set a threshold $\beta$ for $\langle QHR \rangle$, under which the $M_i$ are forced to grow. Conversely, when $\langle QHR \rangle > \beta$ the $M_i$ are forced to decrease. Conversely, $F_2$ and $F_3$ does not allow to set a threshold for $\langle QHR \rangle$. In the following of the article, we focus on $F_2$ and $F_4$ (illustrated in Figure \ref{fig:Fi}), which are representative of the two approaches.

\subsection{Adaptive plan}

\begin{figure}[!hbt]
\centering
\includegraphics[width=12cm]{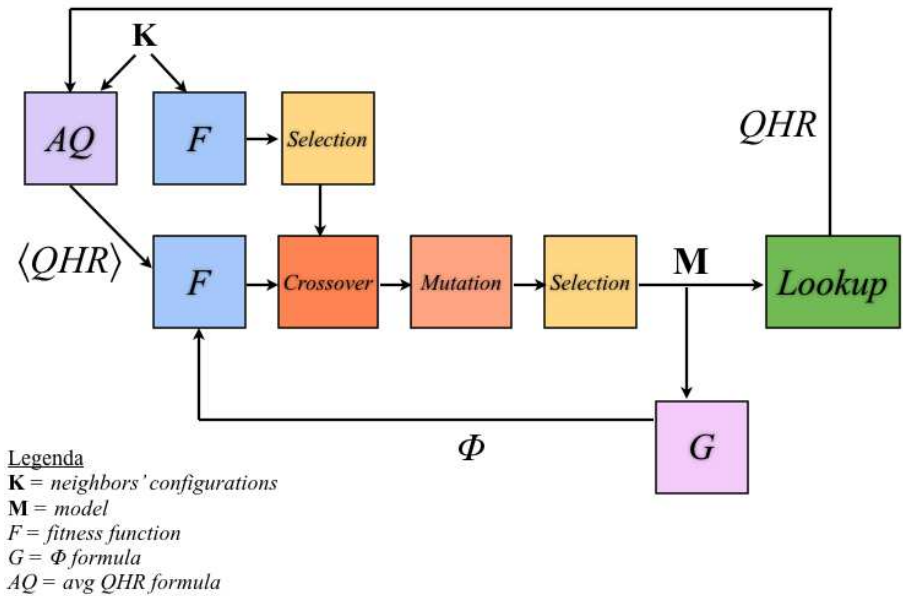}
\caption{Functional architecture of the peer.}
\label{fig:blockDiagram}
\end{figure}

The logical operation of the peer is illustrated by the block diagram in Fig. \ref{fig:blockDiagram}. The resource lookup process is executed in a separate thread with respect to the adaptation process. However, they influence each other.

\begin{algorithm}
\caption{Adaptation}
\label{alg:adaptation}
\begin{algorithmic}[1]
\begin{footnotesize}
\STATE let g = 0
\WHILE{not converged}
    \STATE evaluate the fitness of own model
    \STATE $g = g+1$
    \STATE select neighbor with best fitting model
    \STATE perform cross-over with best neighbor to generate offspring $\mathcal{O}_g = \{\mathbf{O}_{g1}$, $\mathbf{O}_{g2}\}$
    \STATE mutate offspring in $\mathcal{O}_g$ with probability $1 - \langle QHR \rangle$
    \STATE select the new generation $\mathcal{M}_g$ from the previous generation $\mathcal{M}_{g-1}$ and the offspring $\mathcal{O}_g$
\ENDWHILE
\end{footnotesize}
\end{algorithmic}
\end{algorithm}

Algorithm \ref{alg:adaptation} describes the adaptation process which is executed by each peer, with period $T_a$. The "best" neighbor is chosen according to a \textit{fitness-proportionate selection} scheme, where the chance of a neighbor's model to be selected is inversely proportional to its fitness value. Cross-over is always performed, with a randomly-generated crosspoint. Mutation depends on $\langle QHR \rangle$, being highly improbable when the $\langle QHR \rangle$ of the peer and its neighbors tends to $1$. In general, the QHR may be less than $1$ for different reasons. In particular circumstances, no resource may be available even if search messages are deeply and widely propagated. 
Also the chromosome of the current peer or the mutated offspring is then chosen using fitness-proportionate selection. Its consequence is that there is a chance some weaker solutions may survive the selection process. 

This approach recalls Evolvable Agent (EvAg) \cite{Laredo2010}, which is an evolutionary algorithm that structures its population by means of a gossiping protocol. Like in Algorithm \ref{alg:adaptation}, each peer is responsible for one solution, which is obtained by combining those of the first neighbors. For the sake of precision, in Algorithm \ref{alg:adaptation}, the peer's actual solution is also recombined with those of the neighbors.

Algorithm \ref{alg:adaptation} may appear too complicated, for the relatively simple model that is used. Alternatively, each peer may test all the possible configurations on its own, and use the one having best fitness. However, in this way, the time required before a group of linked peers reached the optimal configuration for the local environment would probably be longer. Another possible strategy could be to compare the local model with those of the neighbors, and select the best one, without using crossover and mutation. However, this could lead to non-optimal configurations of the whole network, as the considered problem domain may have a complex fitness landscape, depending on the specific characteristics of the network itself.

\section{Simulations}
\label{simulations}
To evaluate the performance of the proposed ultra-large-scale system, we used a general-purpose discrete event simulation environment, called DEUS, based on Java and XML, released as open source under the GPL license \cite{Amoretti2009b}.
Such a tool has been used instead of classical network simulators (\textit{e.g.}, ns-2 or Opnet), since it is optimized to analyze P2P networks at a higher level. In particular, with DEUS it is possible to simulate highly dynamic overlays (\textit{i.e.} application level) networks, with several hundred thousands nodes, on a single machine --- without the need to simulate also lower network layers (whose effect can be taken into account, in any case, when defining the virtual time scheduling of message propagation events). With respect to PeerSim, a highly reliable simulation tool specifically designed for simulating peer-to-peer networks \cite{Montresor2009}, the approach of DEUS is much simpler, in general: the user just need to extend three classes to define his/her own events, nodes and time-regulating processes. Everything else is managed by the simulation engine.

Each simulated peer is characterized by three kinds of consumable resources: CPU, RAM, and disk space. Their values are randomly generated (with uniform distribution) as multiple of, respectively, $512$ MHz, $256$ MB, and $10$ GB. The maximum amount of resources per peer is $2$ GHz, $1$ GB for the RAM, and $100$ GB for the disk space.

We assumed $M_i$ $\in [1..6]$ ($n=6$), and we set
\begin{itemize}
\item $f_k = M_0 / 6$
\item $T_{max} = M_1$
\item $D_{max} = 2 M_2$
\end{itemize}
to have reasonable values. Regarding $f_k$, that choice is mandatory because it must be in $[0,1]$.

The phenotype $\Phi$ is obtained by means of (\ref{eq:Phi}), with $\phi_0 = 100$, $\phi_1 = 10$, and $\phi_2 = 5$. With these weight values, $f_k$ has more importance than $T_{max}$ and $D_{max}$ in the computation of the fitness value. The average query hit ratio $\langle QHR \rangle$ is given by (\ref{eq:QHR}). To avoid biasing, the initial $QHR$ value for every peer is $0.5$. 

We compared the effects of the fitness functions $F_2$ and $F_4$ (defined in section \ref{casestudy}), that are substantially different because the second one is characterized by a threshold under which $\langle QHR \rangle$ is considered to be bad, while the first one has no explicit constraints.

All the experiments were carried out with a simulated P2P network of $N=10000$ nodes, with a topology constructed without preferential attachment, starting from $N_0$ completely connected nodes, and each other peer being attached to $m \in [1, N_0]$ existing peers, with even probability. All connections are bidirectional, \textit{i.e.} if node $n_a$ has node $n_b$ in its peerview, then $n_b$ has $n_a$ in its peerview. The resulting degree distribution is exponential:
\begin{equation*}
P(k) = (1-e^{- \frac{1}{m}})e^{(1-\frac{k}{m})} \quad \forall k \geq m
\end{equation*}
with expected value
\begin{equation*}
E(k) = \sum_{k=m}^\infty kP(k) = \frac{e^{(1-\frac{1}{m})}}{1-e^{- \frac{1}{m}}} 
\end{equation*}
In these experiments, we used the typical values $m=3$ and $N_0=5$ \cite{Barabasi99}. 
After the initial topology has been constructed, network churn is simulated.
Node departures and joins are modeled by means of two Poisson processes with the same rate $\lambda_1 = 0.14$ s$^{-1}$, for which the average network size remains $N=10000$.

We considered two scenarios: \textit{static load} and \textit{changing load}.
In the former, we simulated $150$ minutes of the life of a network of peers, with resource queries occurring continuously and independently of one another --- according to a Poisson process with parameter $\lambda_2 = 0.028$ s$^{-1}$, \textit{i.e.} $36$ queries every second, each one associated to a randomly chosen node. Each query is a request for a randomly generated amount of resources (no more than $2$ GHz of CPU, $1$ GB of RAM, and $100$ GB of disk space, respectively). Once a resource has been found, it is consumed for a random time interval (according to an exponential distribution with mean value $280$ s --- for which the system's utilization would be $1$, if a provider was found for every request).

In the second scenario (changing load), we still simulated $150$ minutes of the life of a network of peers, but we considered a load on system changing over time. In the first $50$ minutes, almost all resource queries ask for $2$ GHz, $1$ GB of RAM, and $100$ GB of disk space (like in previous experiments). In the following $100$ minutes the load is reduced, with almost all resource queries asking for $256$ MHz, $128$ MB of RAM, and $1$ GB of disk space. 

\begin{figure*}[!hbt]
\centering
\includegraphics[width=8cm]{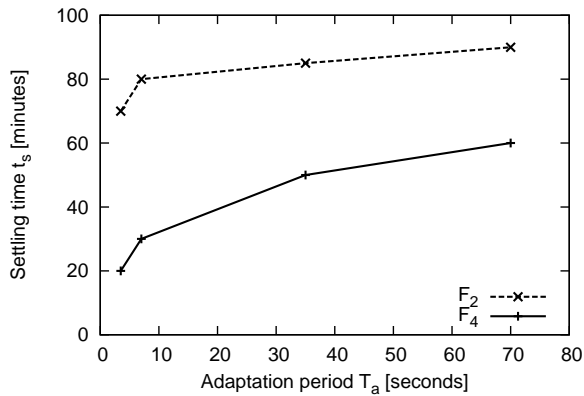}
\caption{Settling time $t_s$ versus adaptation period $T_a$ in the "static load" scenario, comparing fitness functions $F_2$ and $F_4$.}
\label{fig:TsTa}
\end{figure*}

Every node performs adaptation every $T_a$ seconds, for both fixed and noisy networks. All the following results refer to experiments with $T_a = 7$ seconds.
Fig. \ref{fig:TsTa} illustrates how the settling time $t_s$ is affected by the adaptation period, for the static load scenario. Here, the settling time is defined as the time required for the response curve to reach and stay within a range of $5\%$ percentage of the final value. We observe that $F_4$ is always better than $F_2$, for any value of $T_a$. Moreover, it is evident that $T_a = 7$ is a good choice for $F_4$, since lower values would not be not very beneficial for the settling time, while higher values would increase it too much. The exchange of control messages between each node and its neighbors, to support the adaptation process, produces negligible bandwidth consumption, since only the models $\mathbf{M}$ are exchanged (and each node has few neighbors). As already stated before, the adaptation process at the level of each node has a limited computational overhead. However, $T_a$ could be considered as part of the phenotype, and mapped into the model $\mathbf{M}$, in order to be dynamically updated depending on the evaluated performance of the peer. This approach will be evaluated in a future work.

For each scenario and fitness function ($F_2$ and $F_4$), we executed $25$ simulations, each one with a different seed for the random number generator. Reported values in the graphs are the averaged values of such simulations. In each simulation run, the global QHR value is obtained by averaging on all the nodes of the network. The standard deviation also refers to such averaging, which is performed during the simulation, at regular intervals. The same approach has been applied for measuring also the global values of $M_i \forall i \in [0,2]$, $E$, $S$, $C$ and $H$.

\begin{figure*}[!hbt]
\centering
\includegraphics[width=5cm]{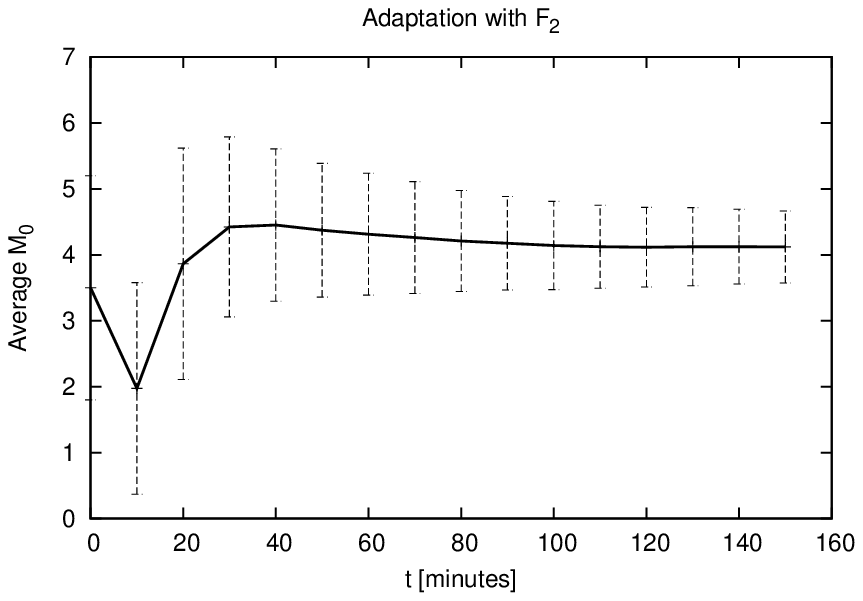}
\includegraphics[width=5cm]{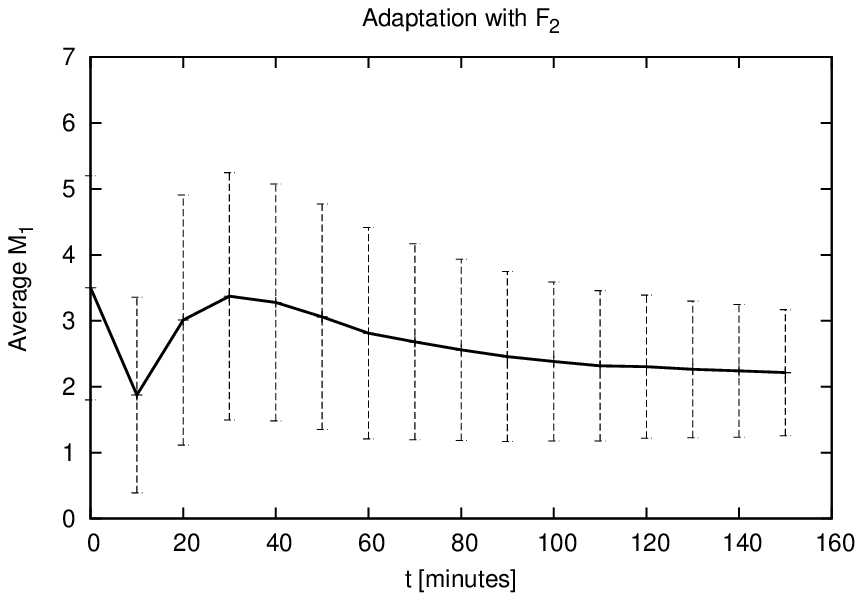}
\includegraphics[width=5cm]{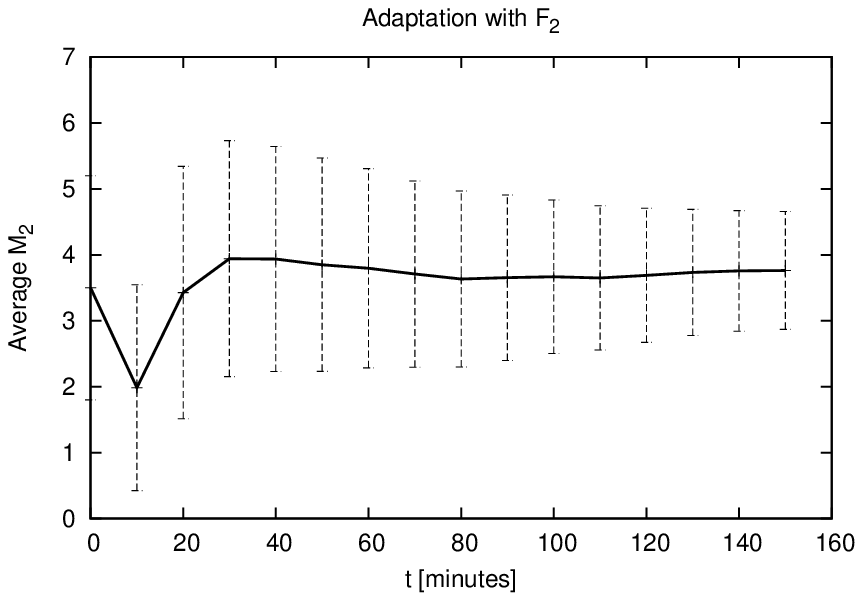}
\includegraphics[width=7cm]{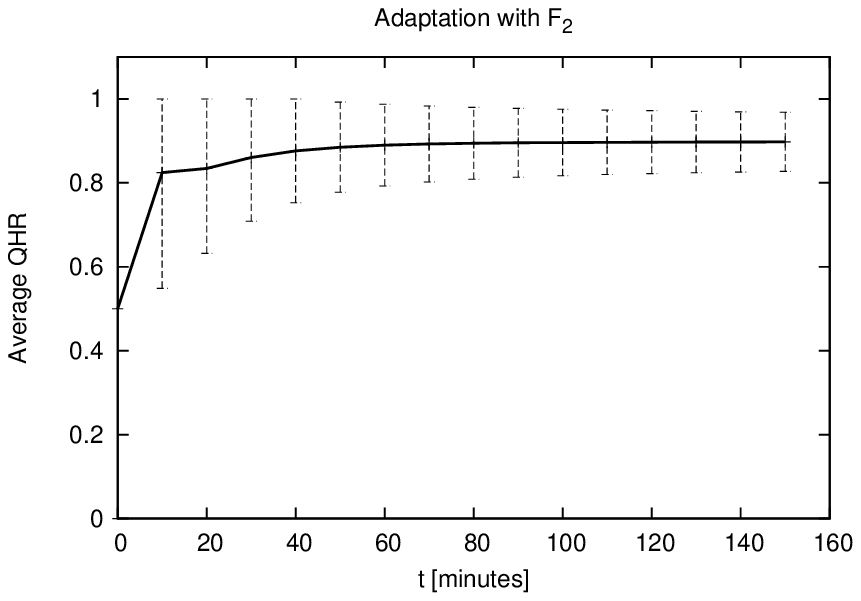}
\caption{Average $M_0$, $M_1$, $M_2$ and $QHR$ over time, for fitness function $F_2$. The load on the system is static.}
\label{fig:miqhrF2-stLoad}
\end{figure*}

\begin{figure*}[!hbt]
\centering
\includegraphics[width=5cm]{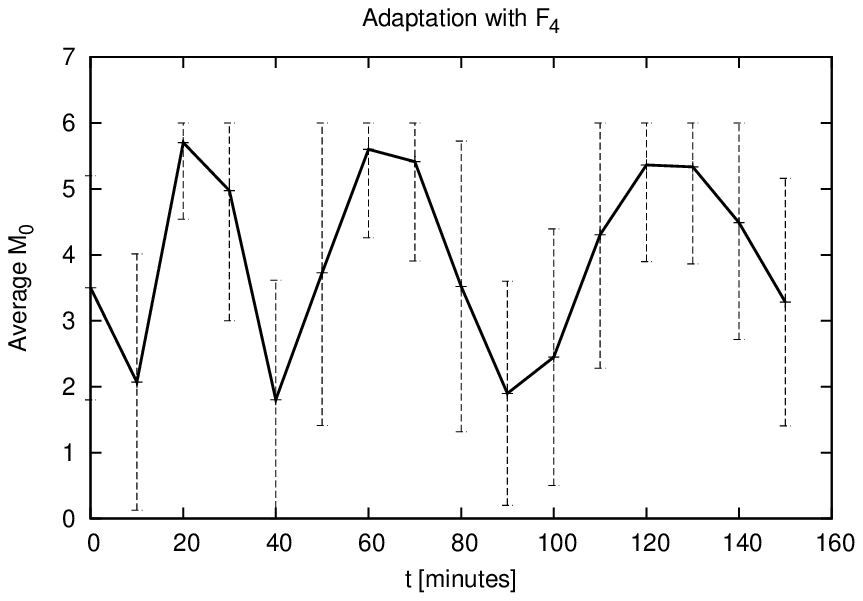}
\includegraphics[width=5cm]{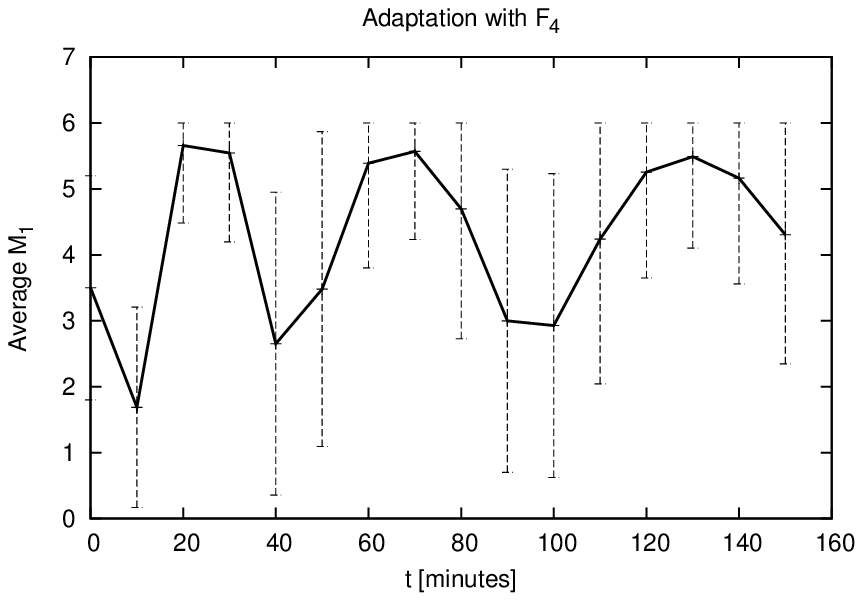}
\includegraphics[width=5cm]{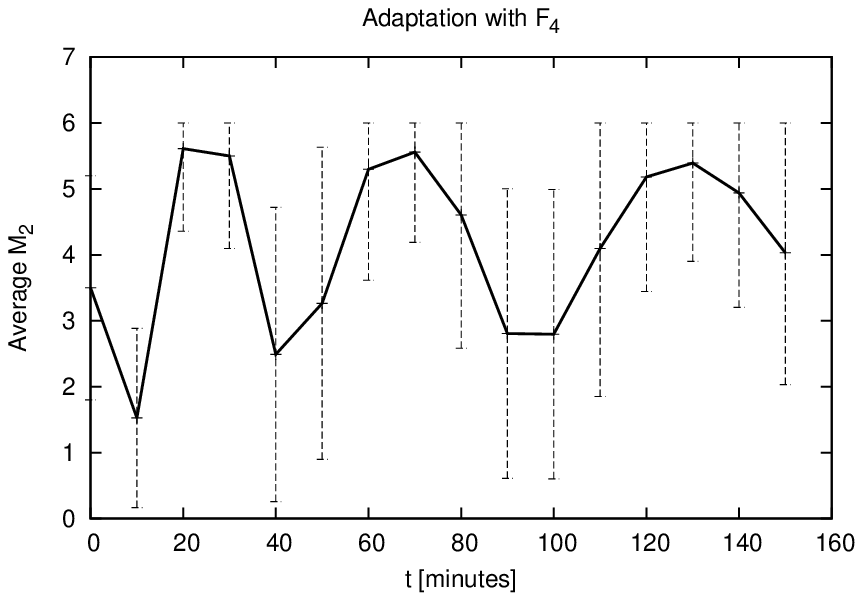}
\includegraphics[width=7cm]{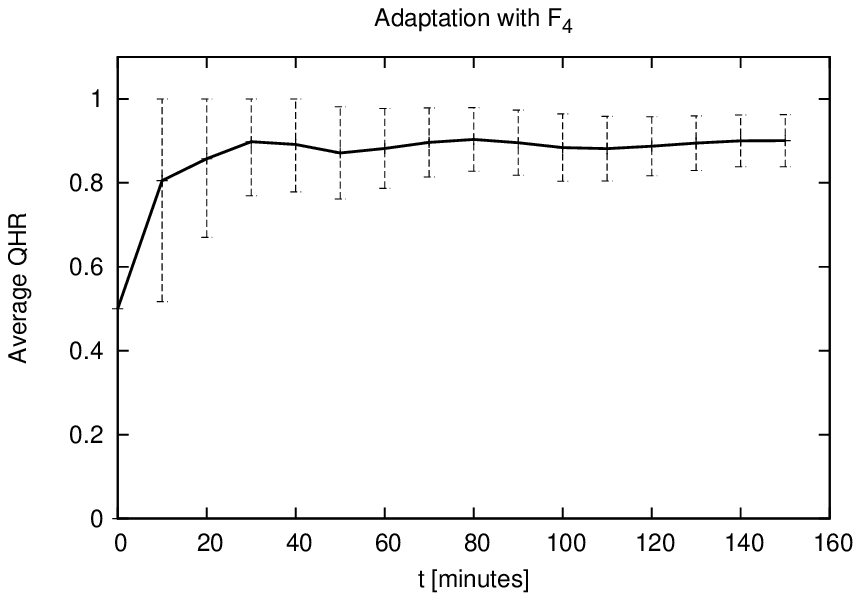}
\caption{Average $M_0$, $M_1$, $M_2$ and $QHR$ over time, for fitness function $F_4$. The load on the system is static.}
\label{fig:miqhrF4-stLoad}
\end{figure*}

Figures \ref{fig:miqhrF2-stLoad} and \ref{fig:miqhrF4-stLoad} illustrate the average evolution of $\mathbf{M}$ and $QHR$ over time, for each considered fitness function, in the static load scenario. 

For the average global $QHR$, we have computed the $95\%$ confidence interval and performed the Wilcoxon rank-sum test. In details, we have considered the two sets of $25$ samples each, representing the global QHR at time $t=150$ minutes, when fitness functions $F_2$ and $F_4$ are used. 
Table 1 shows that, for the scenario with static load, with $F_2$, the values of the global $QHR$ fall in the interval $I95 = [0.8548; 0.9406]$; with $F_4$, they fall in $I95 = [0.8898;0.9115]$. In the same scenario, the Wilcoxon rank-sum test says that the distribution of the global $QHR$, when $F_2$ is used, has the same expected mean value than the one we obtain when $F_4$ is used. The p-value is $0.272 > 0.05$, which means that the null hypothesis 
\begin{quote}
$H_0$: the median value of the distribution of the outputs from $F_2$ equals the median value of the distribution of the outputs from $F_4$
\end{quote}
is confirmed. The second row of the table shows the analysis of the standard deviation of the global $QHR$, which is higher with $F_2$ than with $F_4$. 

\begin{center}
\begin{table*}[h!]
\begin{tabular}{|c|c|c|c|c|c|c|c|}
\hline
\multicolumn{8}{|c|}{Statistical analysis} \\
  \hline
  \textbf{Variable} & \textbf{$\mu_{F_2}$} & \textbf{$\sigma_{F_2}$} & \textbf{$\mu_{F_4}$} & \textbf{$\sigma_{F_4}$} & $I95_{F_2}$ & $I95_{F_4}$ & {\scriptsize $p$-value ($F_2 = F_4$)} \\
  \hline
{\scriptsize   avg global QHR} & {\scriptsize $0.8977$} & {\scriptsize $0.0207$} & {\scriptsize $0.9$} & {\scriptsize $0.0052$} & {\scriptsize $[0.8548;0.9406]$} & {\scriptsize $[0.8898;0.9115]$} & {\scriptsize $0.272$} \\ 
  \hline
  {\scriptsize std dev global QHR} & {\scriptsize $0.0705$ }& {\scriptsize $0.0087$} & {\scriptsize $0.0621$} & {\scriptsize $0.0033$} & {\scriptsize $[0.0523;0.0886]$} & {\scriptsize $[0.0551;0.0691]$} & {\scriptsize $5 \cdot 10^{-4}$} \\
  \hline
\end{tabular}
\caption{Adaptation with static load ($25$ samples $\forall F_i$)}
\end{table*}
\end{center}

\begin{figure*}[!hbt]
\centering
\includegraphics[width=5cm]{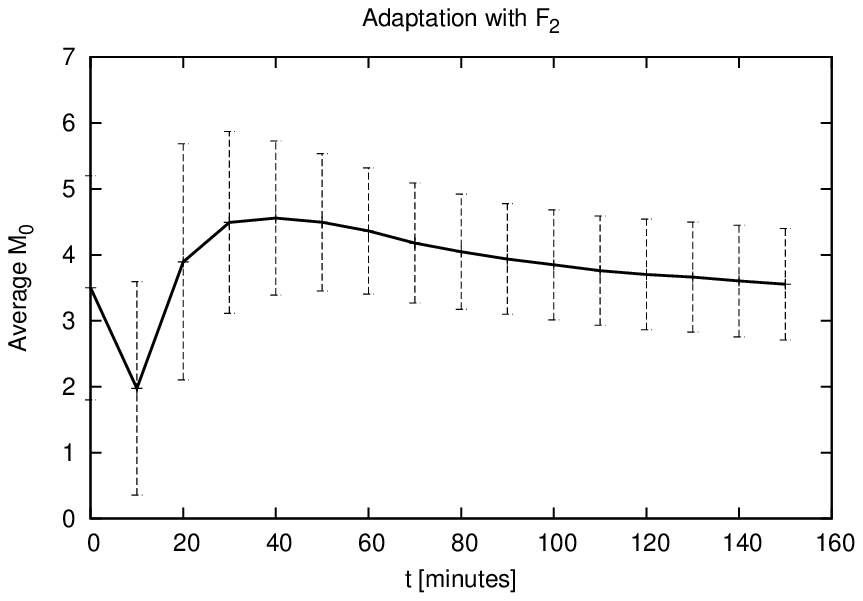}
\includegraphics[width=5cm]{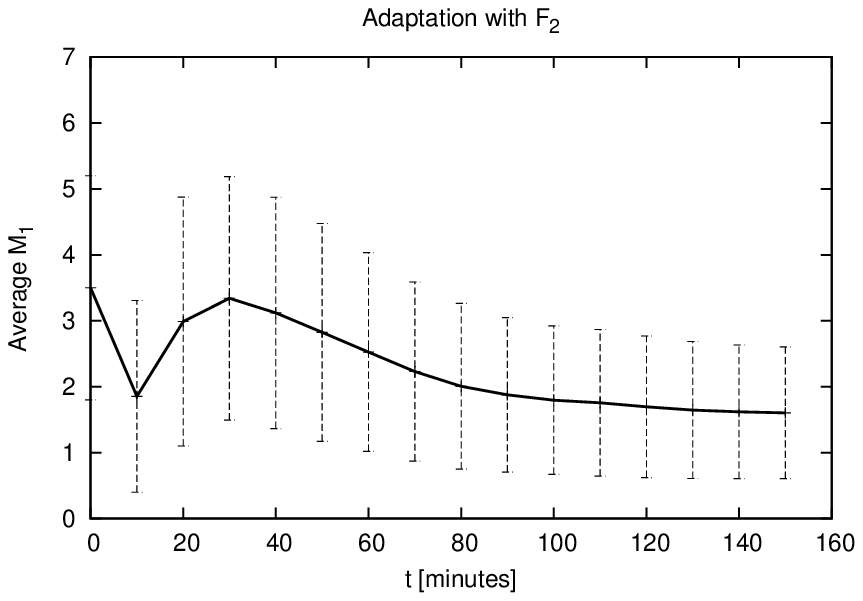}
\includegraphics[width=5cm]{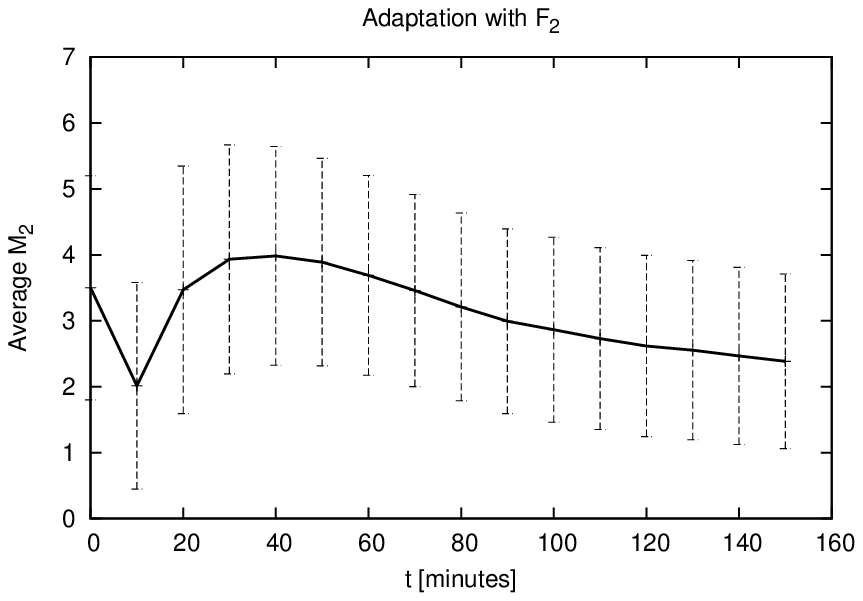}
\includegraphics[width=7cm]{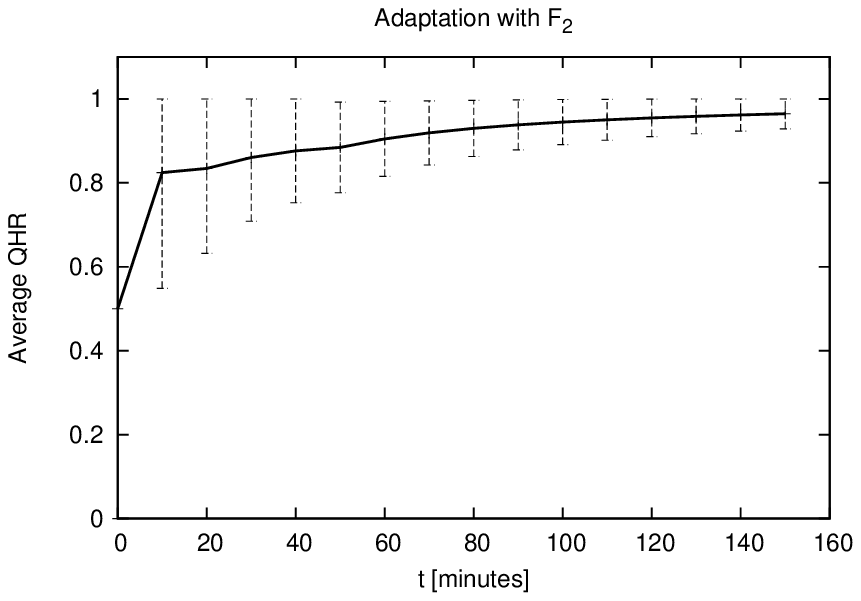}
\caption{Average $M_0$, $M_1$, $M_2$ and $QHR$ over time, for fitness function $F_2$. The load on the system changes at minute $50$.}
\label{fig:miqhrF2-chLoad}
\end{figure*}

\begin{figure*}[!hbt]
\centering
\includegraphics[width=5cm]{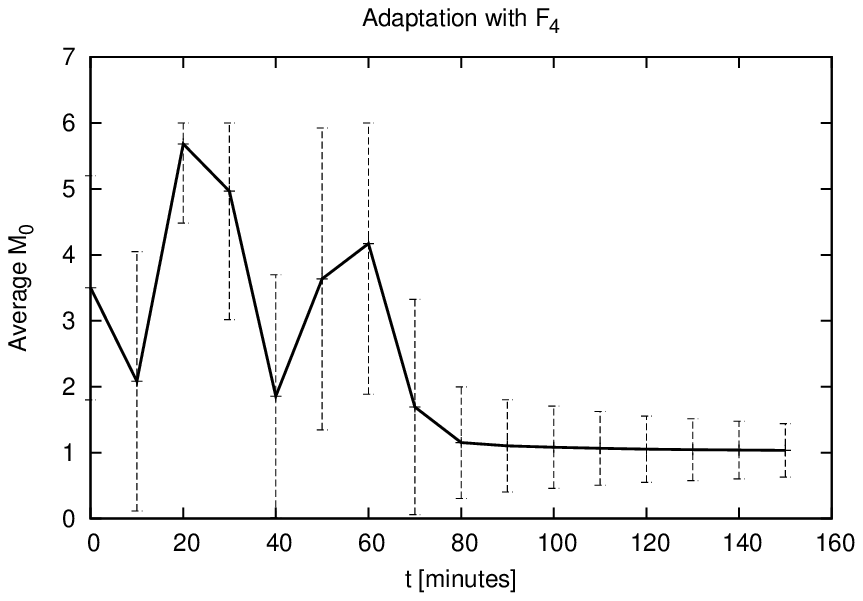}
\includegraphics[width=5cm]{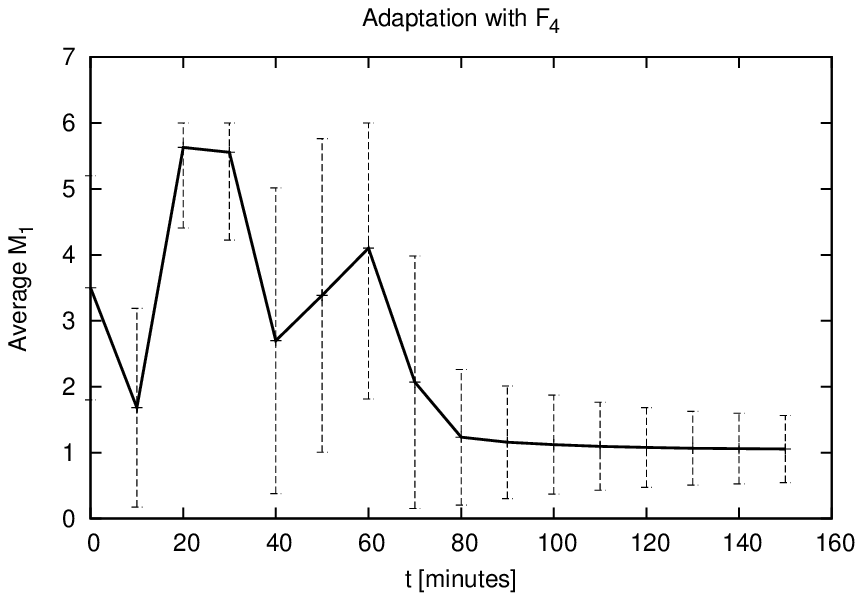}
\includegraphics[width=5cm]{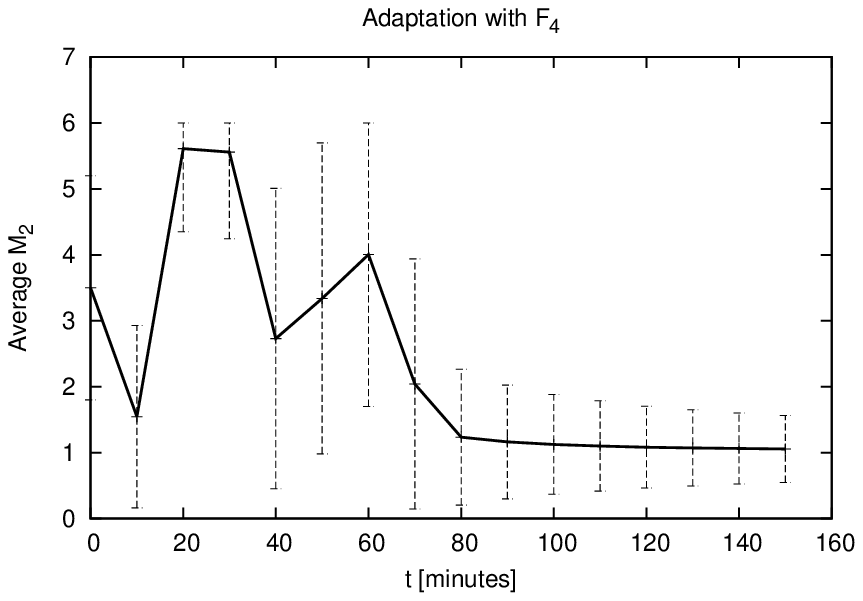}
\includegraphics[width=7cm]{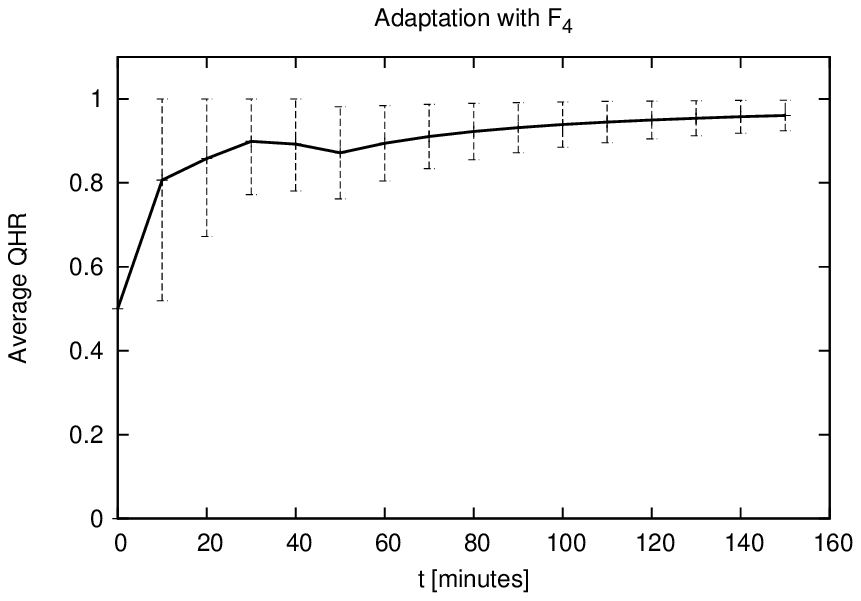}
\caption{Average $M_0$, $M_1$, $M_2$ and $QHR$ over time, for fitness function $F_4$. The load on the system changes at minute $50$.}
\label{fig:miqhrF4-chLoad}
\end{figure*}

The same characterization, for the scenario with changing load, is reported by Fig.\ref{fig:miqhrF2-chLoad} and Fig.\ref{fig:miqhrF4-chLoad}. 

Table 2 shows that in this scenario (dynamic load), the global $QHR$ has higher mean value and lower standard deviation (computed over the $25$ runs) if $F_2$ is used. The fact that $F_2$ leads to better results than $F_4$, over multiple simulation runs, is confirmed by the $I95$ and rank-sum analyses. Once more, we recall that this particular observation refers only to the "stable value" of the global $QHR$, and the performance analysis must be completed by observing the transient phases. It is evident, from figures \ref{fig:miqhrF2-chLoad} and \ref{fig:miqhrF4-chLoad}, that $F_4$ leads to a more responsive system, with respect to $F_2$. This is further discussed below.
\begin{center}
\begin{table*}[h!]
\begin{tabular}{|c|c|c|c|c|c|c|c|}
\hline
\multicolumn{8}{|c|}{Statistical analysis} \\
  \hline
  \textbf{Variable} & \textbf{$\mu_{F_2}$} & \textbf{$\sigma_{F_2}$} & \textbf{$\mu_{F_4}$} & \textbf{$\sigma_{F_4}$} & $I95_{F_2}$ & $I95_{F_4}$ & {\scriptsize $p$-value ($F_2 > F_4$)} \\
  \hline
  {\scriptsize avg global QHR} & {\scriptsize $0.9646$} & {\scriptsize $0.0019$} & {\scriptsize $0.9606$} & {\scriptsize $8 \cdot 10^{-4}$} & {\scriptsize $[0.9606;0.9687]$} & {\scriptsize $[0.9589;0.9623]$} & {\scriptsize $1 \cdot 10^{-8}$} \\ 
  \hline
  {\scriptsize std dev global QHR} & {\scriptsize $0.036$} & {\scriptsize $0.0013$} & {\scriptsize $0.0364$} & {\scriptsize $5 \cdot 10^{-4}$} & {\scriptsize $[0.0332;0.0387]$} & {\scriptsize $[0.0352;0.0376]$} &{\scriptsize  $0.8$} \\
  \hline
\end{tabular}
\caption{Adaptation with changing load ($25$ samples $\forall F_i$)}
\end{table*}
\end{center}

In general, when $QHR$ decreases for most peers, the system reacts in a way that chromosomes with higher gene values survive and proliferate. Both the considered fitness functions make the genes $M_i$ change over time (with an initial transient phase), to maximize the $QHR$. 
Because of node departures and arrivals, the genes are characterized by high standard deviation. The reason is that joining and leaving nodes introduce random chromosomes that produce the same effect as mutation.
With $F_4$ the standard deviation of the genes is particularly high, because of the firm threshold $\beta$ which characterizes $F_4$: if $\langle QHR \rangle$ oscillates around $\beta$, the genes do not reach a stable configuration. However, this produces a positive effect on the global $QHR$. 

In the second scenario, with changing load, both the considered fitness functions make the genes $M_i$ increase in the first period (high load), and decrease in the second period (reduced load). We notice that $F_4$ leads to a more prompt reaction, \textit{i.e.} gene values decrease almost immediately when the environmental conditions change. Such a reactive behavior is highly appreciated, considering the large size of the system.

\begin{figure*}[!hbt]
\centering
\includegraphics[width=7cm]{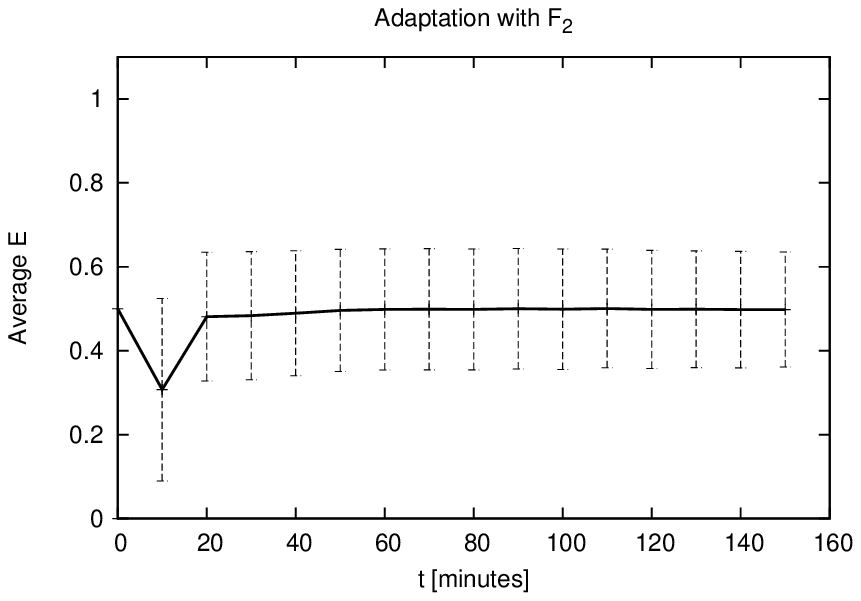}
\includegraphics[width=7cm]{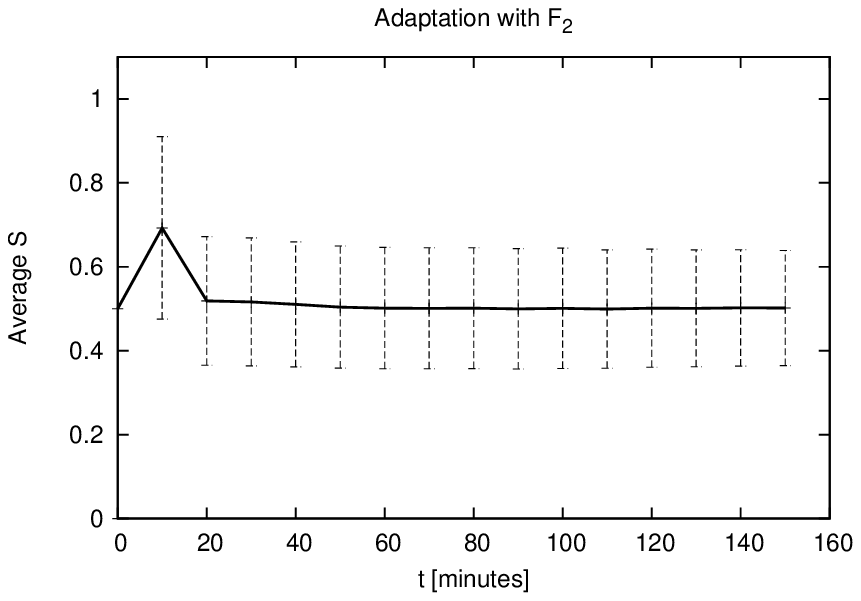}
\includegraphics[width=7cm]{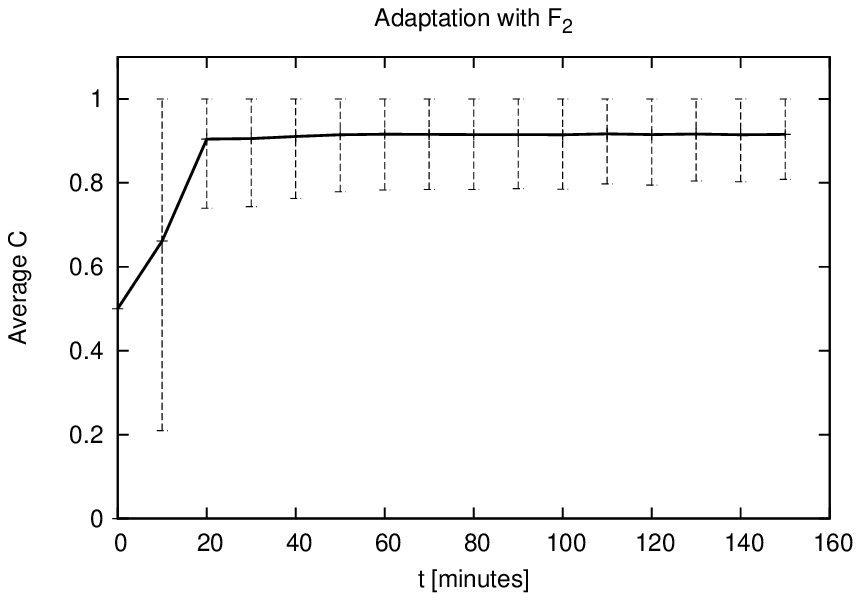}
\includegraphics[width=7cm]{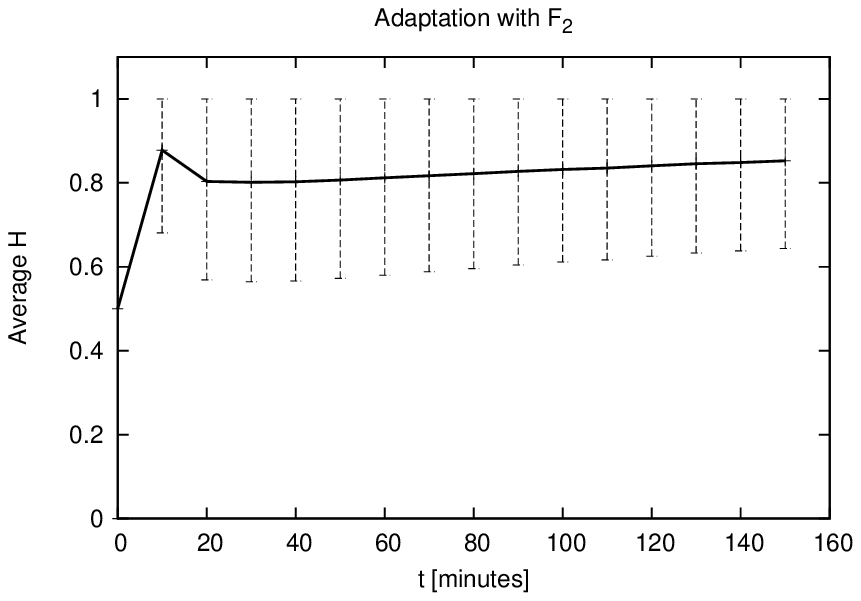}
\caption{Average $E$, $S$, $C$ and $H$ over time, for fitness function $F_2$. The load on the system is static.}
\label{fig:eschF2-stLoad}
\end{figure*}

\begin{figure*}[!hbt]
\centering
\includegraphics[width=7cm]{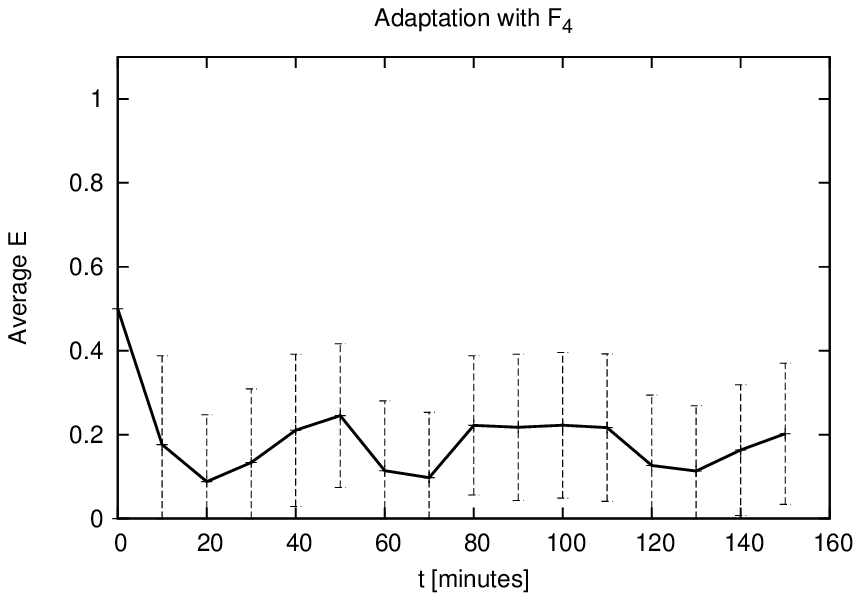}
\includegraphics[width=7cm]{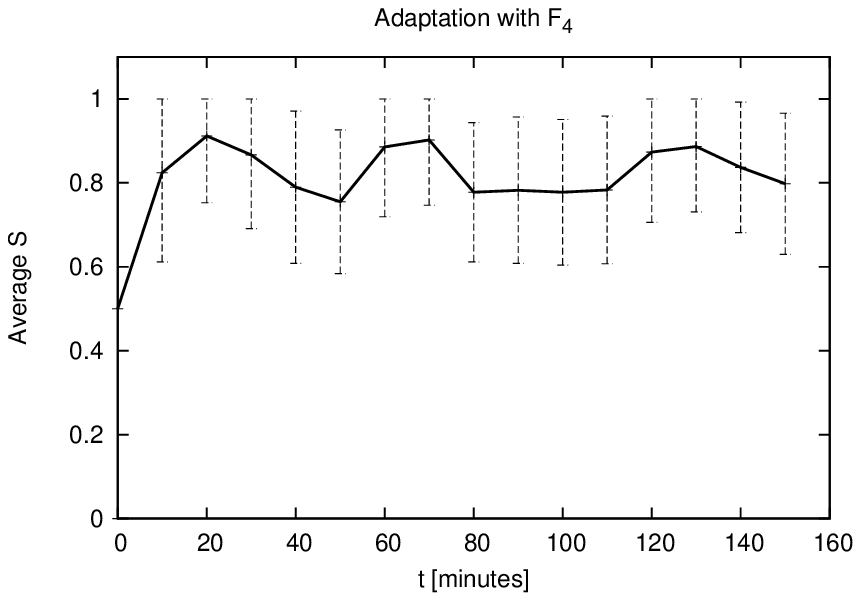}
\includegraphics[width=7cm]{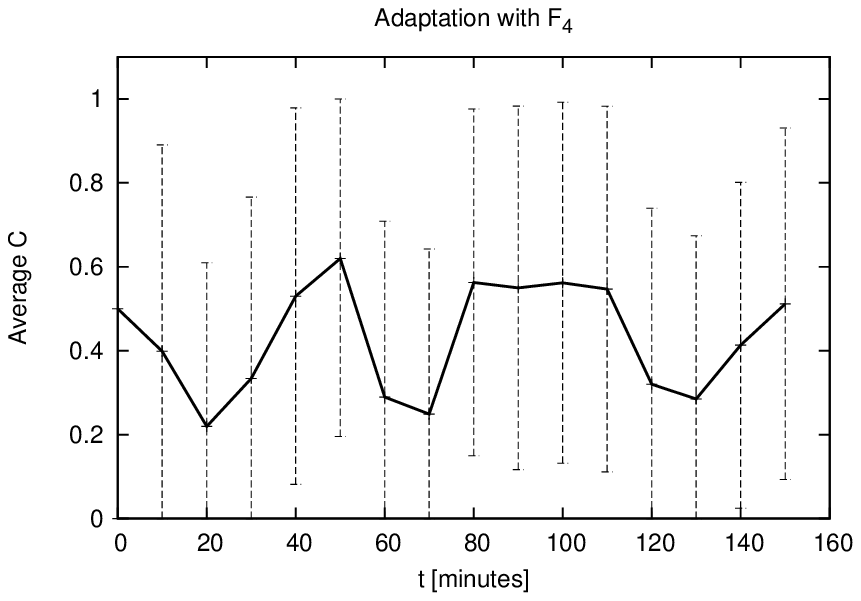}
\includegraphics[width=7cm]{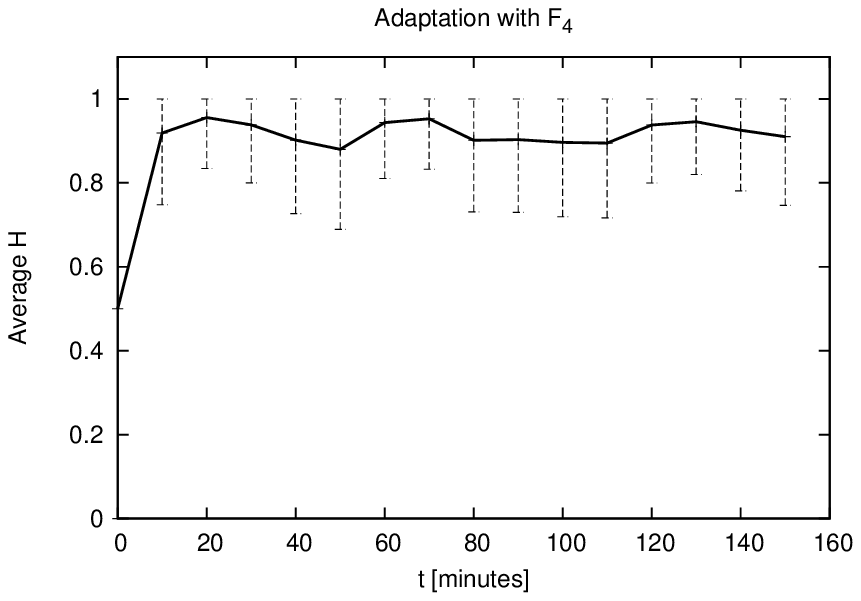}
\caption{Average $E$, $S$, $C$ and $H$ over time, for fitness function $F_4$. The load on the system is static.}
\label{fig:eschF4-stLoad}
\end{figure*}

\begin{figure*}[!hbt]
\centering
\includegraphics[width=7cm]{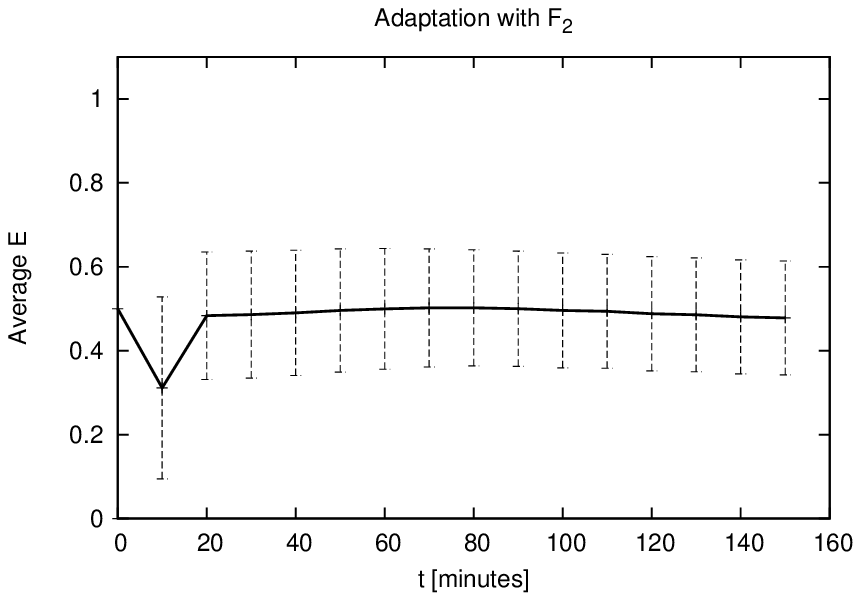}
\includegraphics[width=7cm]{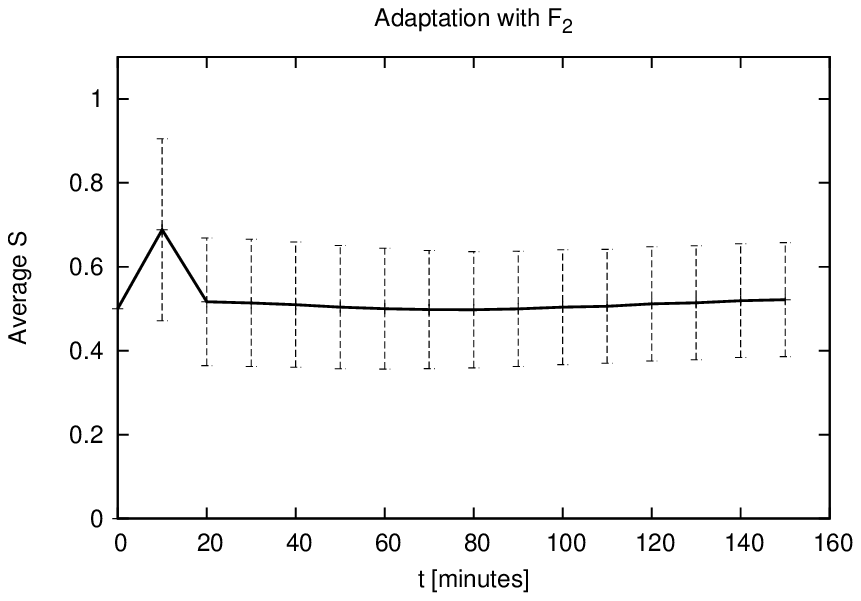}
\includegraphics[width=7cm]{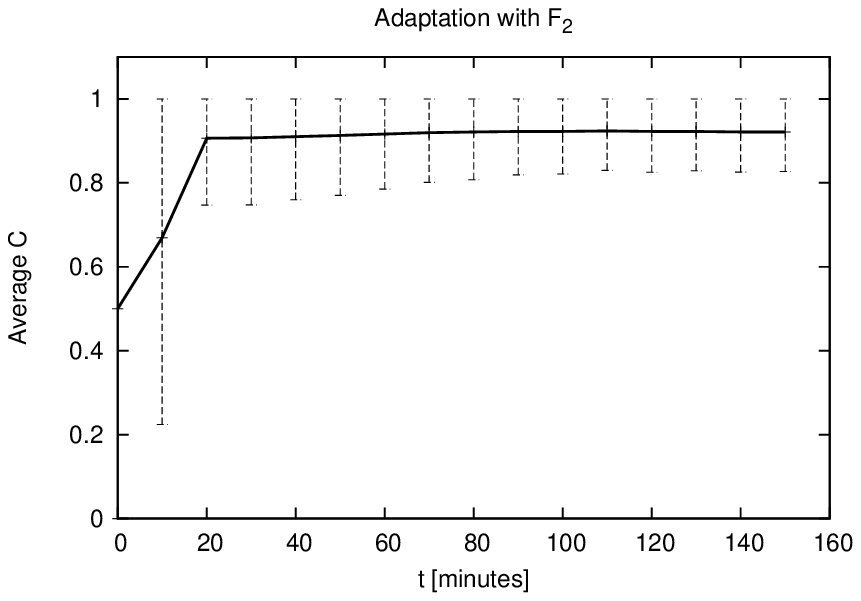}
\includegraphics[width=7cm]{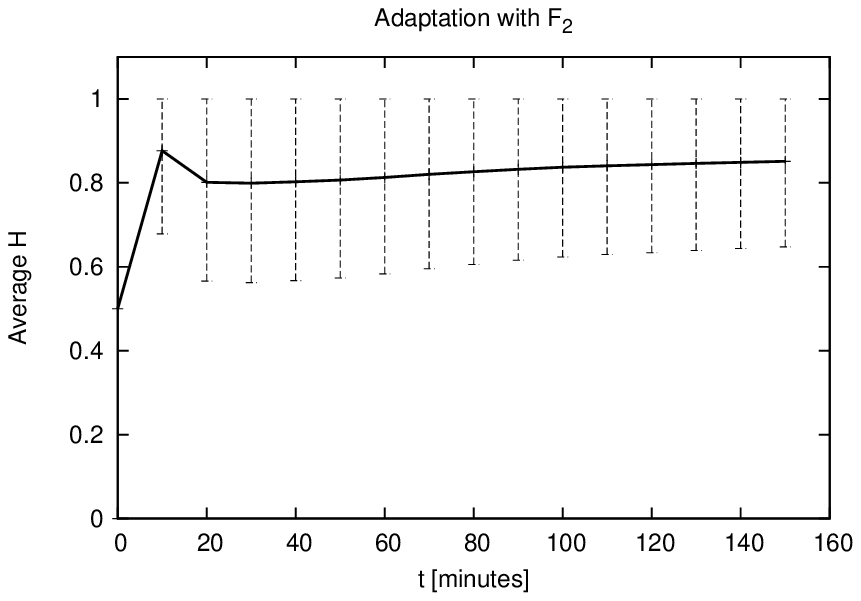}
\caption{Average $E$, $S$, $C$ and $H$ over time, for fitness function $F_2$. The load on the system changes at minute $50$.}
\label{fig:eschF2-chLoad}
\end{figure*}

\begin{figure*}[!hbt]
\centering
\includegraphics[width=7cm]{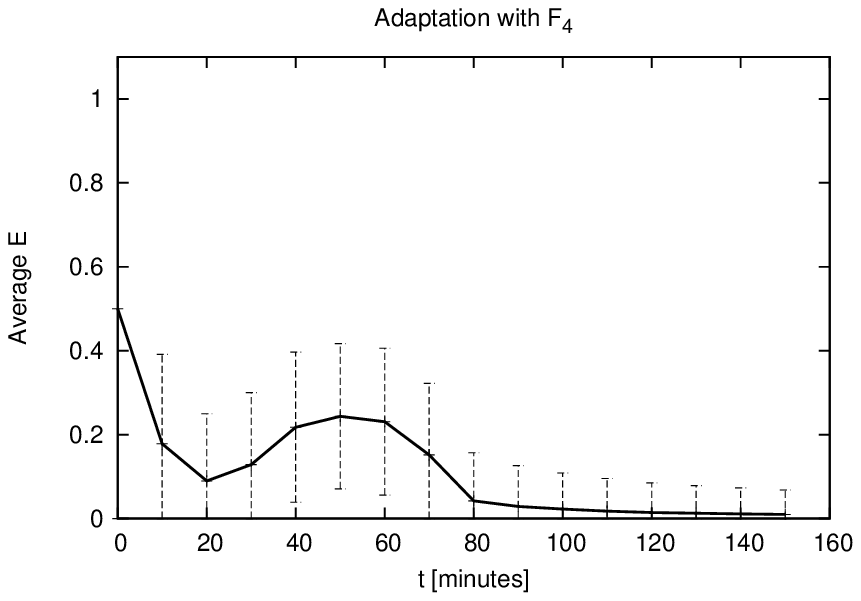}
\includegraphics[width=7cm]{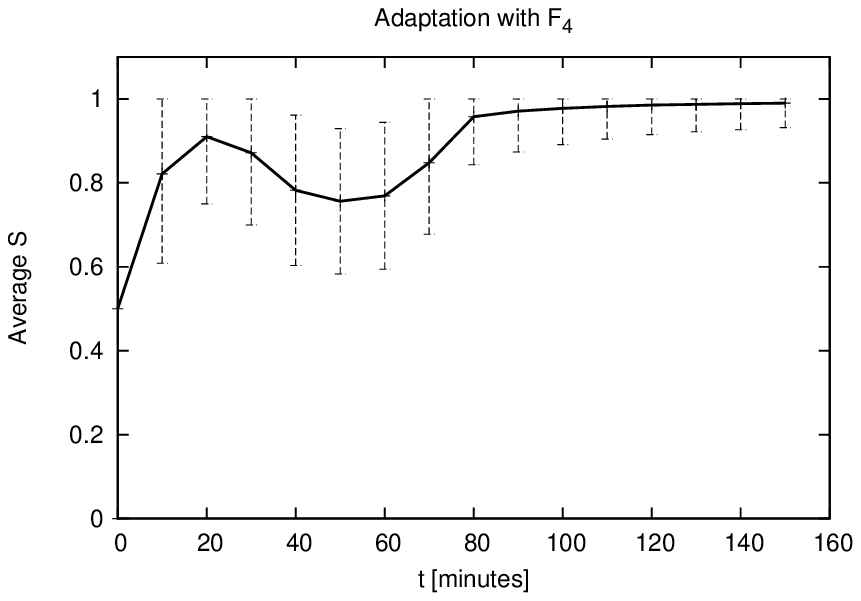}
\includegraphics[width=7cm]{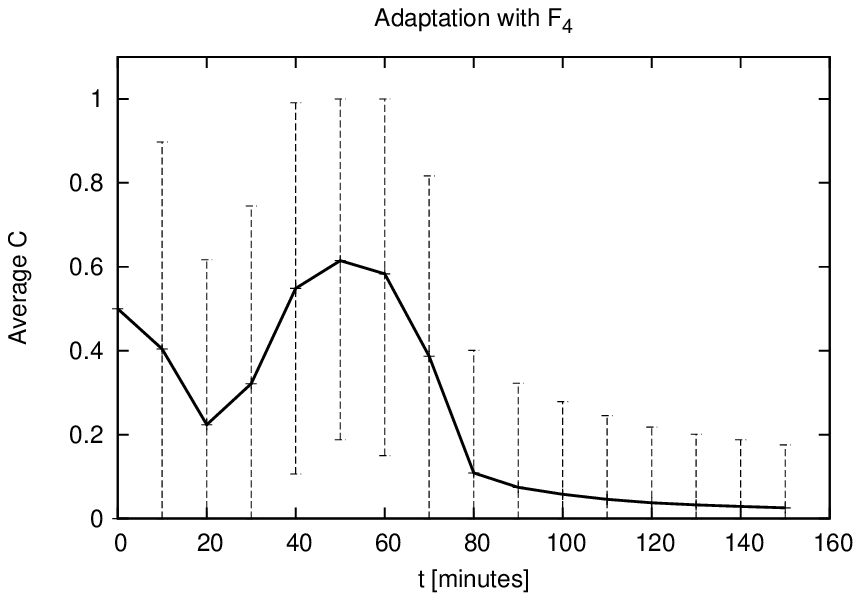}
\includegraphics[width=7cm]{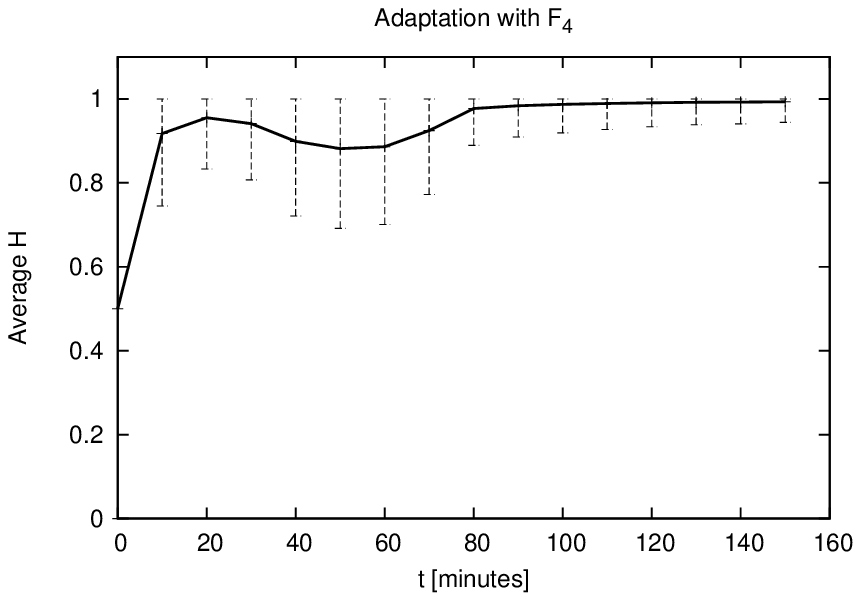}
\caption{Average $E$, $S$, $C$ and $H$ over time, for fitness function $F_4$. The load on the system changes at minute $50$.}
\label{fig:eschF4-chLoad}
\end{figure*}

We measured also the evolution over time of the information $I$ associated to the current genotype $\mathbf{M}$, according to the procedure illustrated in section \ref{methodology}. Figures \ref{fig:eschF2-stLoad} to \ref{fig:eschF4-chLoad} are related to the case of $W=1$.

For the static load scenario, $F_2$ produces balanced change in time, reflected in medium $E$ and $S$ and a high $C$.  $F_4$ offers less change on average, as seen by a higher $S$ and $H$, lower $E$, and medium $C$. Still, the rate of information change changes considerably in time, leading to a variance in the measures, especially high for $C$. 

For the changing load scenario, $F_2$ performs in a  similar fashion: it is not able to adapt to the reduction of the demand. Conversely, $F_4$ does adapt at the demand change at minute 50, stabilizing at minute 80. This is seen in maximal $S$ and $H$ and minimal $E$ and $C$, indicating that there is no change in the nodes. This is because of the threshold used in $F_4$.

Higher $S$ and $H$ imply static behavior, which might be desired after the load change, in the second scenario. High $E$ implies change, while a high $C$ implies a balance between $S$ and $E$, \textit{i.e.} adaptivity. A high $C$ is desirable when changes are being made to the load, not when there are no changes. 

\section{Conclusion}
\label{conclusion}

We proposed the use of measures based on information theory to study the performance of ULS systems. These measures are general enough to be used with any proposed ULS, enabling comparison between different approaches.

We presented simulation results of an ULS computing system with two different types of genetic adaptation while measuring their complexity, emergence, self-organization and homeostasis. The main result is that less "aggressive" adaptive plans lead to a more stable system, but the optimal performance may not be achieved.

Future work will follow two main directions. Firstly, with respect to the use case proposed in this paper, we will study the impact of parameter variations and we will consider also other workload profiles. Then, we will generalize the analysis to the case of systems with components that dynamically change their structure, by removing or adding components.

%% References


\bibliographystyle{elsarticle-num}



% List: Number the references (numbers in square brackets) in the list in the order in which they appear in the text.

%Examples: 
%Reference to a journal publication: 
%[1] J. van der Geer, J.A.J. Hanraads, R.A. Lupton, The art of writing a scientific article, J. Sci. Commun. 163 (2010) 51Ð59. 
%Reference to a book: 
%[2] W. Strunk Jr., E.B. White, The Elements of Style, fourth ed., Longman, New York, 2000. 
%Reference to a chapter in an edited book: 
%[3] G.R. Mettam, L.B. Adams, How to prepare an electronic version of your article, in: B.S. Jones, R.Z. Smith (Eds.), Introduction to the Electronic Age, E-Publishing Inc., New York, 2009, pp. 281Ð304.

\end{document}